\documentclass[letterpaper, 10 pt, conference]{ieeeconf}  

\IEEEoverridecommandlockouts                              
\makeatletter
\let\NAT@parse\undefined
\makeatother

\usepackage[numbers,sectionbib,sort&compress]{natbib}

\usepackage{todonotes}
\usepackage{graphicx} 
\usepackage{times} 
\usepackage{amsmath} 
\usepackage{amssymb}  
\usepackage{todonotes}
\usepackage{verbatim} 
\usepackage{subfig}
\usepackage[hyphens]{url}
\usepackage[ruled,vlined]{algorithm2e}
\usepackage{float}
\usepackage{tabulary}
\usepackage{tabularx}
\usepackage{lipsum,amsmath,multicol}
\usepackage{tensor}
\usepackage{multirow}
\usepackage{booktabs}
\usepackage{todonotes}

\usepackage{hyperref}
\hypersetup{colorlinks,breaklinks,
linkcolor=[rgb]{0.5,0.,0.},
citecolor=[rgb]{0.000,0.427,0.173},
urlcolor=[rgb]{0.031,0.318,0.612}}

\def\fps@figure{htp}
\def\fps@table{htp}

\newcommand{\bi}{\begin{itemize}}
\newcommand{\ei}{\end{itemize}}

\newcommand{\bfig}{\begin{figure}}
\newcommand{\efig}{\end{figure}}

\newcommand{\benum}{\begin{enumerate}}
\newcommand{\eenum}{\end{enumerate}}

\newcommand{\be}{\begin{equation}}
\newcommand{\ee}{\end{equation}}

\newcommand{\ba}{\begin{eqnarray}}
\newcommand{\ea}{\end{eqnarray}}

\newcommand{\unit}[1]{\mbox{$\rm \,#1$}}

\usepackage{color}

\definecolor{CommentRed}{rgb}{0.7,0,0}
\definecolor{CommentBlue}{rgb}{0,0,0.7}
\definecolor{CommentDG}{rgb}{0,0.6,0}

\makeatletter
\newenvironment{myalign*}{%
  \setlength{\mathindent}{0pt}%
  \setlength{\abovedisplayskip}{-\baselineskip}%
  \setlength{\abovedisplayshortskip}{\abovedisplayskip}%
  \start@align\@ne\st@rredtrue\m@ne
}%
{\endalign}
\makeatother

\title{\LARGE \bf
weedNet: Dense Semantic Weed Classification Using Multispectral Images and MAV for Smart Farming
}

\author{$\text{Inkyu Sa}^{1}$, $\text{Zetao Chen}^{2}$, $\text{Marija Popovi\'{c}}^{1}$, $\text{Raghav Khanna}^{1}$, $\text{Frank Liebisch}^{3}$, $\text{Juan Nieto}^{1}$, $\text{Roland Siegwart}^{1}$\thanks{${}^1$ Autonomous Systems Lab., ${}^2$ Vision for Robotics Lab., Department of Mechanical and Process Engineering, ${}^3$ Crop Science, Department of Environmental Systems Science, ETH Zurich, Switzerland. \texttt{inkyu.sa@mavt.ethz.ch}}
}

\begin{document}

\maketitle

\thispagestyle{empty}
\pagestyle{empty}


\begin{abstract}
Selective weed treatment is a critical step in autonomous crop management as related to crop health and yield. However, a key challenge is reliable, and accurate weed detection to minimize damage to surrounding plants. In this paper, we present an approach for dense semantic weed classification with multispectral images collected by a micro aerial vehicle (MAV). We use the recently developed encoder-decoder cascaded Convolutional Neural Network (CNN), Segnet, that infers dense semantic classes while allowing any number of input image channels and class balancing with our sugar beet and weed datasets. To obtain training datasets, we established an experimental field with varying herbicide levels resulting in field plots containing only either crop or weed, enabling us to use the Normalized Difference Vegetation Index (NDVI) as a distinguishable feature for automatic ground truth generation. We train 6 models with different numbers of input channels and condition (fine-tune) it to achieve $\sim$ 0.8 F1-score and 0.78 Area Under the Curve (AUC) classification metrics. For model deployment, an embedded GPU system (Jetson TX2) is tested for MAV integration. Dataset used in this paper is released to support the community and future work.
\end{abstract}

\section{INTRODUCTION}
\label{sec:intro}

To sustain a growing worldwide population with sufficient farm produce, new smart farming methods are required to increase or maintain crop yield while minimizing environmental impact.
Precision agriculture techniques achieve this by spatially surveying key indicators of crop health and applying treatment, e.g. herbicides, pesticides, and fertilizers, only to relevant areas.
Here, robotic systems can be often used as flexible, cost-efficient platforms replacing laborious manual procedures.

\begin{figure}
\begin{center}
\includegraphics[width=\columnwidth]{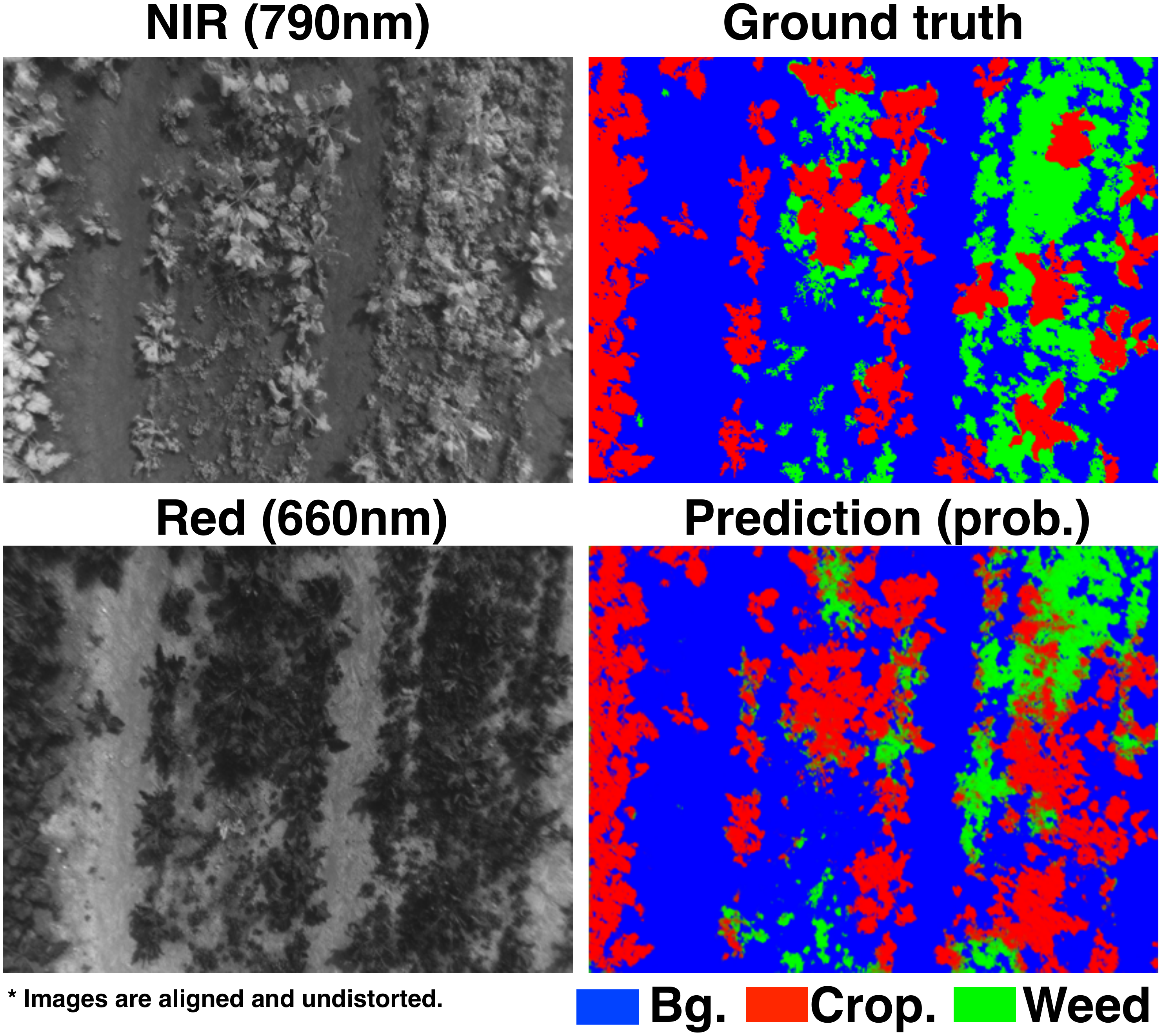}
\end{center}
    \vspace{-3mm}
    \caption{NIR (top-left) and Red channel (bottom-left) input images with ground truth (top-right).
    Bottom-right is the probability output from our dense semantic segmentation framework.}
    \label{fig:front}
        \vspace{-5mm}
\end{figure}

Specifically, weed treatment is a critical step in autonomous farming as it directly associates with crop health and yield~\citep{slaughter2008autonomous}. Reliable, and precise weed detection is a key requirement for effective treatment as it enables subsequent processes, e.g. selective stamping, spot spraying, and mechanical tillage, while minimizing damage to surrounding vegetation. However, accurate weed detection presents several challenges. Traditional object-based classification approaches are likely to fail due to unclear crop-weed boundaries, as exemplified in Fig.~\ref{fig:front}. This aspect also impedes manual data labeling which is required for supervised learning algorithms.

To address these issues, we employ a state-of-the-art dense (pixel-wise) Convolutional Neural Network (CNN) for segmentation.
We chose this CNN because it can predict relatively faster than other algorithms while maintaining competitive performance. The network utilizes a modified VGG16~\citep{Simonyan:2014aa} architecture (i.e., dropping last two fully-connected layers) as an encoder and a decoder is formed with upsampling layers that are counterparts for each convolutional layer in the encoder.

Our CNN encodes visual data as low-dimensional features capturing semantic information, and decodes them back to higher dimensions by up-sampling. To deal with intensive manual labeling tasks, we maneuver a micro aerial vehicle (MAV) with a downward-facing multispectral camera over a designed field with varying herbicide levels applied to different crop field areas. To this end, we obtain 3 types of multispectral image datasets which contain (i) only crop, (ii) only weed, and (iii) both crop and weed. For (i) and (ii), we can easily create ground truth by extracting Normalized Difference Vegetation Index (NDVI) indicating vegetation cover. Unfortunately, we could not avoid manual labeling for (iii), which took about 60\unit{mins} per image (with 30 testing images). Given these training and testing datasets, we train 6 different models with varying input channels and training conditions (i.e., fine-tune) to see the impact of these variances. The trained model is deployed on an embedded GPU computer that can be mounted on a small-scale MAV. 

The contributions of this system paper are:
\begin{itemize}
\item Release of pixel-wise labelled (ground-truth) sugar beet/weed datasets collected from a controlled field experiment\footnote{\label{foot:video_data} Available at: \url{https://goo.gl/UK2pZq}},
\item A study on crop/weed classification using dense semantic segmentation with varying multispectral input channels.
\end{itemize}

Since the dense semantic segmentation framework predicts the probability for each pixel, the outputs can be easily used by high-level path planning algorithms, e.g. monitoring-~\citep{Popovic2017IROS} or exploration-based~\citep{bircher2016receding}, for informed data collection and complete autonomy on the farm.

The remainder of this paper is structured as follows.
Section~\ref{sec:background} presents the state-of-the-art on pixel-wise semantic segmentation, vegetation detection using a MAV, and CNN with multispectral images. Section~\ref{sec:methodologies} describes how the training/testing dataset is obtained and details of our model training procedure. We present our experimental results in Section \ref{sec:results}, and conclude the paper in Section \ref{sec:conclusion}.
\section{RELATED WORK}\label{sec:background}
This section briefly reviews the state-of-the-art in deep segmentation models, general methods of detecting vegetation, and segmentation based on multispectral images. 

\subsection{Pixel-wise Segmentation using Deep Neural Network}
The aim of the image segmentation task is to infer a human-readable class label for each image pixel, which is an important and challenging task. The most successful approaches in recent years rely on CNN.  Early CNN-based methods perform segmentation in a two-step pipeline, which first generates region proposals and then classifies each proposal to a pre-defined category ~\citep{girshick2014rich,hariharan2014simultaneous}. Recently, fully Convolutional Neural Networks (FCNNs) have become a popular choice in image segmentation, due to their rich feature representation and end-to-end training~\citep{chen2016deeplab,dai2015boxsup}. However, these FCNN-based methods usually have a limitation of low-resolution prediction, due to the sequential max-pooling and down-sampling operation. SegNet~\cite{badrinarayanan2015segnet}, the module that our weedNet is based on, is a recently proposed pixel-wise segmentation module that carefully addresses this issue. It has an encoder and a corresponding decoder network. The encoder network learns to compress the image into a lower-resolution feature representation, while the decoder network learns to up-sample the encoder feature maps to full input resolution for per-pixel segmentation. The encoder and decoder networks are trained simultaneously and end-to-end.

\subsection{MAV-based Crop-Weed Detection}
For smart farming applications, it is becoming increasingly important to accurately estimate the type and distribution of the vegetation on the field. There are several existing approaches which exploit different types of features and machine learning algorithms to detect vegetation~\citep{Liebisch2017,Sa:2016aa,Haug2014}. \citet{torres2015automatic} and \citet{khanna2015beyond} investigate the use of NDVI and Excess Green Index (EGI) to automatically detect vegetation from soil background. Comparatively, \citet{guo2013illumination} exploit spectral features from RGB images and decision tress to separate vegetation. A deeper level of smart farming is an automatic interpretation of the detected vegetation into classes of crop and weed. \citet{perez2015semi} utilize multispectral image pixel values as well as crop row geometric information to generate features for classifying image patches into valuable crop, weed, and soil. Similarly, \citet{pena2013weed} exploit spatial and spectral characteristics to first extract image patches, and then use the geometric information of the detected crop rows to distinguish crops and weeds. In~\citep{Lottes:2017aa}, visual features as well as geometric information of the detected vegetation are employed to classify the detected vegetation into crops and weeds using Random Forest algorithms~\citep{breiman2001random}. 
All the above-mentioned approaches either directly operate on raw pixels or rely on a fixed set of handcrafted features and learning algorithms. 

However, in the presence of large data, recent developments of deep learning have shown end-to-end learning approaches outperforms traditional hand-crafted feature learning~\citep{girshick2014rich}. Inspired by this, \citet{mortensen2016semantic} proposed a CNN-based semantic segmentation approach to classify different types of crops and estimate their individual amount of biomass. Compared to their approach which operates on RGB images, the approach in this paper extract information from multispectral images using a different state-of-the-art per-pixel segmentation model. 

\subsection{Applications using Multispectral Images}
Multi-spectral images provide the possibility to create vegetation specific indices based on radiance ratios which are more robust under varying lighting conditions and therefore are widely explored for autonomous agriculture robotics~\citep{mccool2016visual,lucieer2014hyperuas,khanna2017field,hung2013orchard}. In~\citep{bac2013robust}, images captured from a six band multi-spectral camera were exploited to segment different parts of sweet pepper plants. Similarly, in~\citep{pena2013weed}, the authors managed to compute weed maps in maize fields from multispectral images. In~\citep{garcia2015sugar}, multispectral images were used to separate sugar beets from a thistle. In our study, we apply a CNN-based pixel-wise segmentation model directly on the multispectral inputs to accurately segment crops from weeds.

As shown above, there are intensive research interests in agricultural robotics domain employing state-of-the-art CNN, multispectral images, and MAVs for rapid field scouting. Particularly, precise and fast weed detection is the key front-end module that can lead subsequent treatments to success. To our best knowledge, this work is the first trial applying a real-time ($\approx$2\unit{Hz}) CNN-based dense semantic segmentation using multispectral images taken from a MAV to agricultural robotics application while maintaining applicable performance. Additionally, we release our dataset utilized in this paper to support the community and future work since there are insufficient public available weed dataset~\citep{Haug:2014aa}.
\section{METHODOLOGIES}
\label{sec:methodologies}
In this section, we present our approaches to dataset acquisition
and pre-processing through image alignment and NDVI extraction
before outlining our model training method.

\subsection{Dataset Acquisition}
Dense manual annotation of crop/weed species is a challenging task, as shown in 
Fig.~\ref{fig:front}.
Unlike urban street (e.g.~\cite{Geiger2013IJRR} and~\cite{Cordts2016Cityscapes}) 
or indoor scenes \cite{song2015sun} that can easily be understood and inferred intuitively,
plant boundaries in our datasets are difficult to distinguish and may require 
domain-specific knowledge or experience.
They also require finer-selection tools (e.g., pixel-level selection and 
zoom-in/out) rather than polygon-based 
outline annotation with multiple control points.
Therefore, it may difficult to use coarse outsourced services~\cite{Sorokin:2008aa}.

\begin{figure}
\begin{center}
\includegraphics[width=\columnwidth]{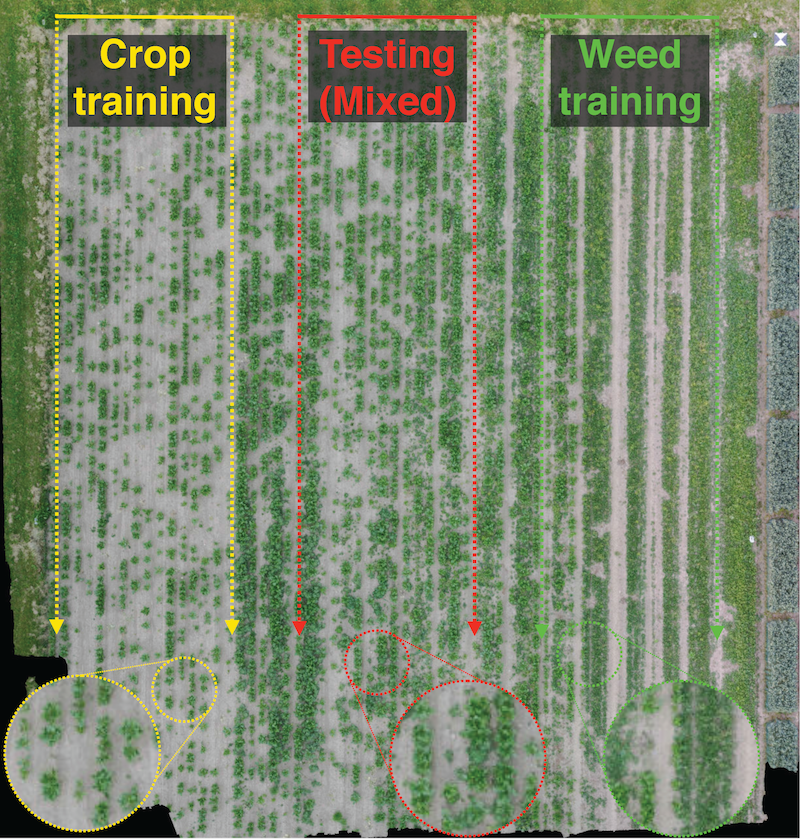}
\end{center}
    \vspace{-4mm}
    \caption{Aerial view of our controlled field with a varying herbicide levels.
    The maximum amount of herbicide is applied to the left crop training rows (yellow),
    and no herbicide is utilized for the right weed training rows (green). The middle shows mixed variants due to medium herbicide usage (red).}
    \label{fig:weedFieldTopView}
        \vspace{-3mm}
\end{figure}

To address this issue, we designed the 40\unit{m}$\times$40\unit{m}
weed test field shown in  Fig.~\ref{fig:weedFieldTopView}.
We applied different levels of herbicide to the field; \unit{max}, \unit{mid}, 
and \unit{min}, corresponding to left (yellow), mid (red), and right 
(green) respectively. Therefore, as expected,  images from the left-to-right field patches contain crop-only, crop/weed, and weed-only, respectively.
We then applied basic automated image processing techniques to extract 
NDVI from the left and right patch images, as described in the following sub-section.
The crop-weed images could only be annotated manually
following advice from crop science experts.
This process took on average about 60\unit{mins}/image.
As shown in Table~\ref{tbl:dataset}, we annotated 132, 243, and 90 multispectral images of crops, weeds, and crop-weed mixtures.
Each training  image/test image consisted of near-infrared (NIR, 790\unit{nm}), Red channel 
(660\unit{nm}), and NDVI imagery.

\begin{table}
  \caption{Image datasets for training and testing}
      \vspace{-2mm}
  \begin{center}
    \begin{tabular}{ccccc}    
    \multirow{2}{*}{}
    (NIR+Red+NDVI) & Crop & Weed & Crop-weed & Num. multispec \\  \midrule \midrule
    Training  & 132 &  243    &---        & 375               \\  \midrule
    Testing  & ---    &  ---          & 90 &90             \\  \midrule
    Altitude (\unit{m})  & 2    &  2          & 2 &    ---         \\
    \end{tabular}
    \end{center}
  \label{tbl:dataset}
      \vspace{-6mm}
\end{table}%

\begin{figure}
\begin{center}
\includegraphics[width=\columnwidth]{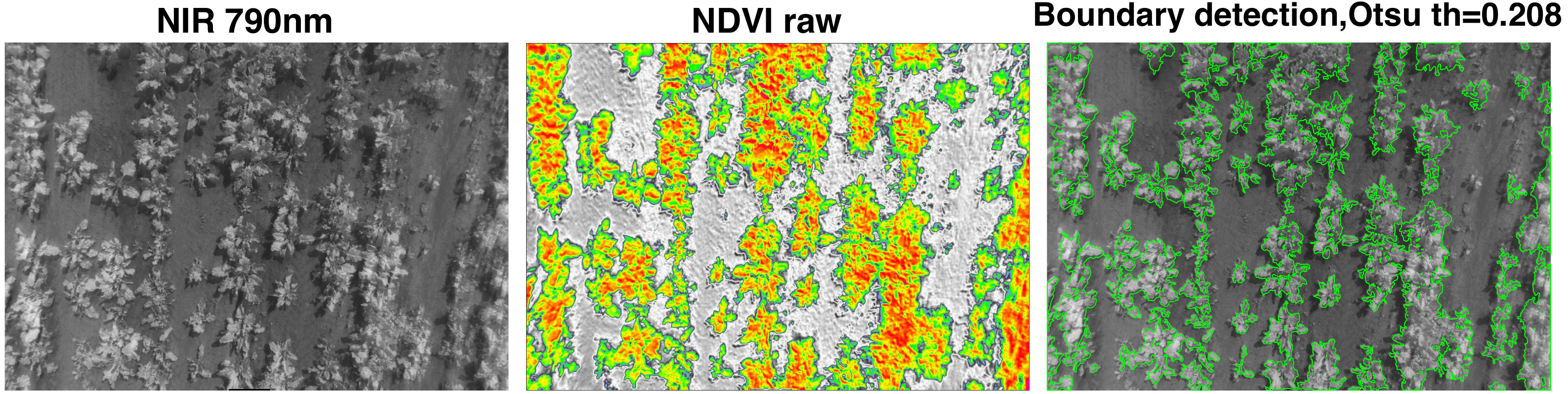}
\end{center}
    \vspace{-4mm}
    \caption{An input crop image (left), with NDVI is extracted using NIR+Red channels (mid.).
    An auto-threshold boundary detection method outputs clear edges between vegetation and others (shadows, soil, and gravel) (right).}
    \label{fig:preproc}
        \vspace{-6mm}
\end{figure}

\subsection{Data Pre-processing}
For image acquisition, we use a Sequoia multispectral sensor that features four narrow band global shutter imagers (1.2\unit{MP}), and one rolling shutter RGB camera (16\unit{MP}).
From corresponding NIR and Red images, we extract the NDVI, given by
$\text{NDVI}=\frac{(\text{NIR}-\text{Red})}{(\text{NIR}+\text{Red})}$.
This vegetation index clearly indicates the difference between soil and plant, as exemplified in
Fig.~\ref{fig:preproc} (NDVI raw).

\subsubsection{Image alignment}
To calculate indices, we performed basic image processing for the NIR and Red 
images through image undistortion, estimation of geometric transformation $\in 
\text{SE3}$ using image correlation, and cropping. Note that the processing 
time for these procedures is negligible since these transformations need to be computed only once for cameras attached rigidly with respect to each other. 
It is also worth mentioning that we could not align the other image channels,
e.g. Green and Red Edge, in the same way due to the lack of similarities.
Furthermore, it is difficult to match them correctly without estimating the depth of each pixel accurately.
Therefore, our method assumes that the 
camera baseline is much smaller than the distance from the ground and 
camera ($\sim$ two orders of magnitude). We only account for camera intrinsics,
and do not apply radiometric and atmospheric corrections.

\subsubsection{NDVI extraction}\label{subsec:ndvi}
We then applied a Gaussian blur to the aligned images (threshold=1.2),
followed by a sharpening procedure to remove fine responses (e.g., shadows, small debris). 
An intensity histogram clustering algorithm, Otsu's method~\cite{otsu1979threshold}, is used for threshold selection on the 
resultant image and blob detection is finally executed with the minimum of 300 
connected pixels. Fig.~\ref{fig:preproc} shows the detected 
boundary of a crop image and each class is labeled as follows \{bg, crop, 
weed\}=\{0,1,2\}.

\subsection{Dense Semantic Segmentation Framework}
\begin{figure}
\begin{center}
\includegraphics[width=\columnwidth]{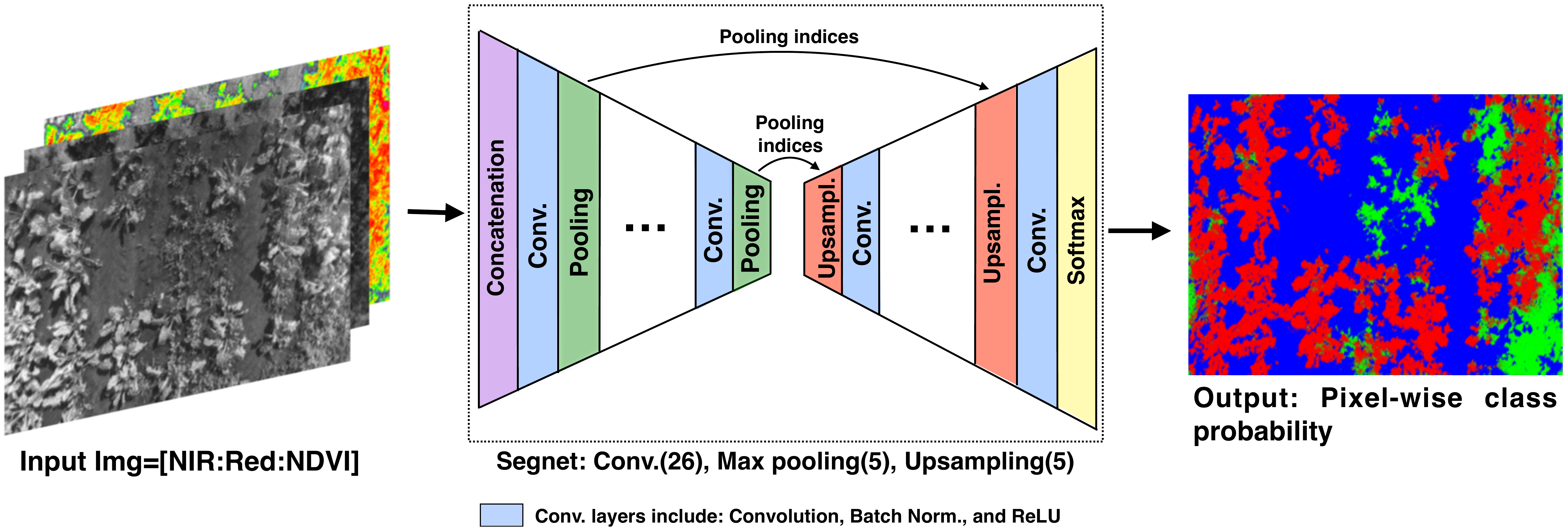}
\end{center}
    \vspace{-3mm}
    \caption{An encoder-decoder cascaded dense semantic segmentation framework~\cite{badrinarayanan2015segnet}. It has 26 convolution layers followed by ReLU activation and 5 max-pooling for the encoder (first half) and 5 up-sampling for the decoder (second half) linked with pooling indices. The first concatenation layer allows for any number of input channels. The output depicts the probability of 3 classes.}
    \label{fig:segnet}
        \vspace{-3mm}
\end{figure}

The annotated images are fed into SegNet, a state-of-the art dense segmentation framework shown in 
Fig.~\ref{fig:segnet}. We retain the original network architecture (i.e., 
VGG16 without fully-connected layers and additionally upsample layers for each 
counterpart for max-pooling)~\cite{badrinarayanan2015segnet} and here
only highlight the modifications performed.

Firstly, the frequency of appearance (FoA) for each class is adapted based on our 
training dataset for better class balancing~\cite{eigen2015predicting}. This is 
used to weigh each class inside the neural network loss function and requires careful tuning.
For example, as the \texttt{weed} class appears less frequently than \texttt{bg} and \texttt{crop},
its FoA is lower in comparison. If a false-positive or false-negative is detected in weed classification (i.e., a pixel is 
incorrectly classified as \texttt{weed}), then the classifier is penalized more than for the
other classes. A class weight can be written as:
    \vspace{-2mm}
\begin{align}
w_{\mbox{\tiny C}}=&\frac{\widetilde{FoA(c)}}{FoA(c)}\\
FoA(c)=&\frac{I_{\mbox{\tiny C}}^{\mbox{\tiny Total}}}{I_{\mbox{\tiny 
C}}^{\mbox{\tiny j}}}
\label{eq:pfh}
\end{align}
where $\widetilde{FoA(c)}$ is the median of $FoA(c)$, $I_{\mbox{\tiny 
C}}^{\mbox{\tiny Total}}$ is the total number of pixels in class $c$, and 
$I_{\mbox{\tiny C}}^{\mbox{\tiny j}}$ is the number of pixels in the $j$th image 
where class $c$ appears, with $j\in{\{1,2,3...N\}}$ as the image sequence number.

Secondly, we implemented a simple input/output layer that reads images and outputs
them to the subsequent concatenation layer. This allows us to feed any 
number of input images to the network,
which is useful for hyperspectral image processing~\cite{Khanna:2017aa}.


\section{EXPERIMENTAL RESULTS}\label{sec:results}
In this section, we present our experimental setup, followed by a qualitative and quantitative evaluation of our proposed approach. We also demonstrate a preliminary performance evaluation of our model deployed on an embedded computer.
\subsection{Experimental Setup}
As shown in Fig.~\ref{fig:weedFieldTopView}, we cultivated a 40\unit{m}$\times$40\unit{m} test sugar beet field with varying herbicide levels applied for automated ground-truth acquisition following the procedures in Section \ref{subsec:ndvi}. 

\begin{figure}
\begin{center} \includegraphics[width=0.8\columnwidth]{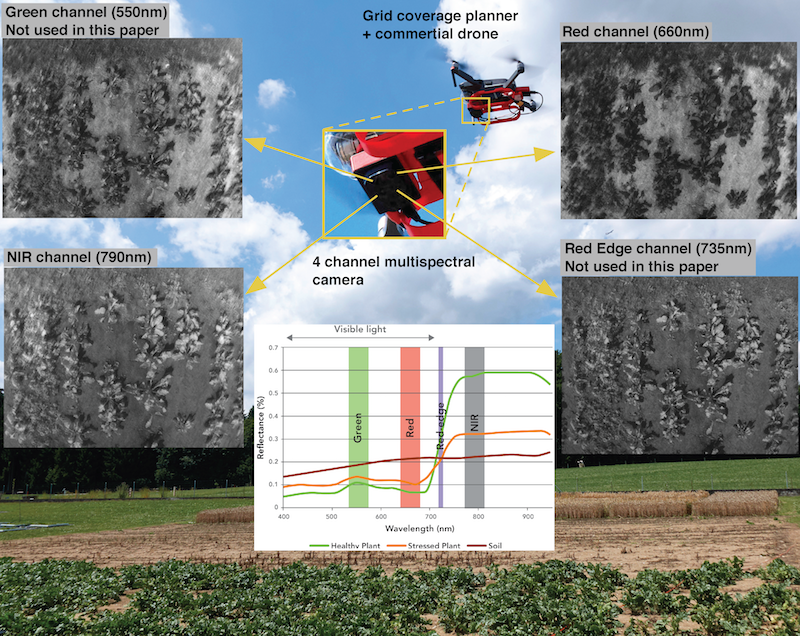}
\end{center}
        \vspace{-4mm}
       \caption{Data collection setup. A DJI Mavic flies over our experimental field with a 4-band multispectral camera. In this paper, we consider only NIR and Red channels due to difficulties in image registration of other bands. The graph illustrates the reflectance of each band for healthy and sick plants, and soil. (image courtesy of Micasense\protect \footnotemark).}
    \label{fig:Mavic}
        \vspace{-5mm}
\end{figure}

A downward-facing Sequoia multispectral camera is mounted on a commercial MAV, DJI Mavic, recording datasets at 1\unit{Hz} (Fig~\ref{fig:Mavic}). The MAV is manually controlled in position-hold mode assisted by GPS and internal stereo-vision system at 2\unit{m} height. It can fly around 15$\sim$17\unit{mins} with  an additional payload of 274.5\unit{g}, including an extra battery pack, Sequoia camera, radiation sensor, a DC-converter, and 3D-printed mounting gear. Once a dataset is acquired, the information is transferred manually to a ground station.
\footnotetext{https://goo.gl/pveB6D}
For model training, we use NVIDIA's Titan X GPU module on a desktop computer and a Tegra TX2 embedded GPU module with an Orbitty carrier board for model inference.

\begin{figure}
\begin{center}
\includegraphics[width=0.8\columnwidth]{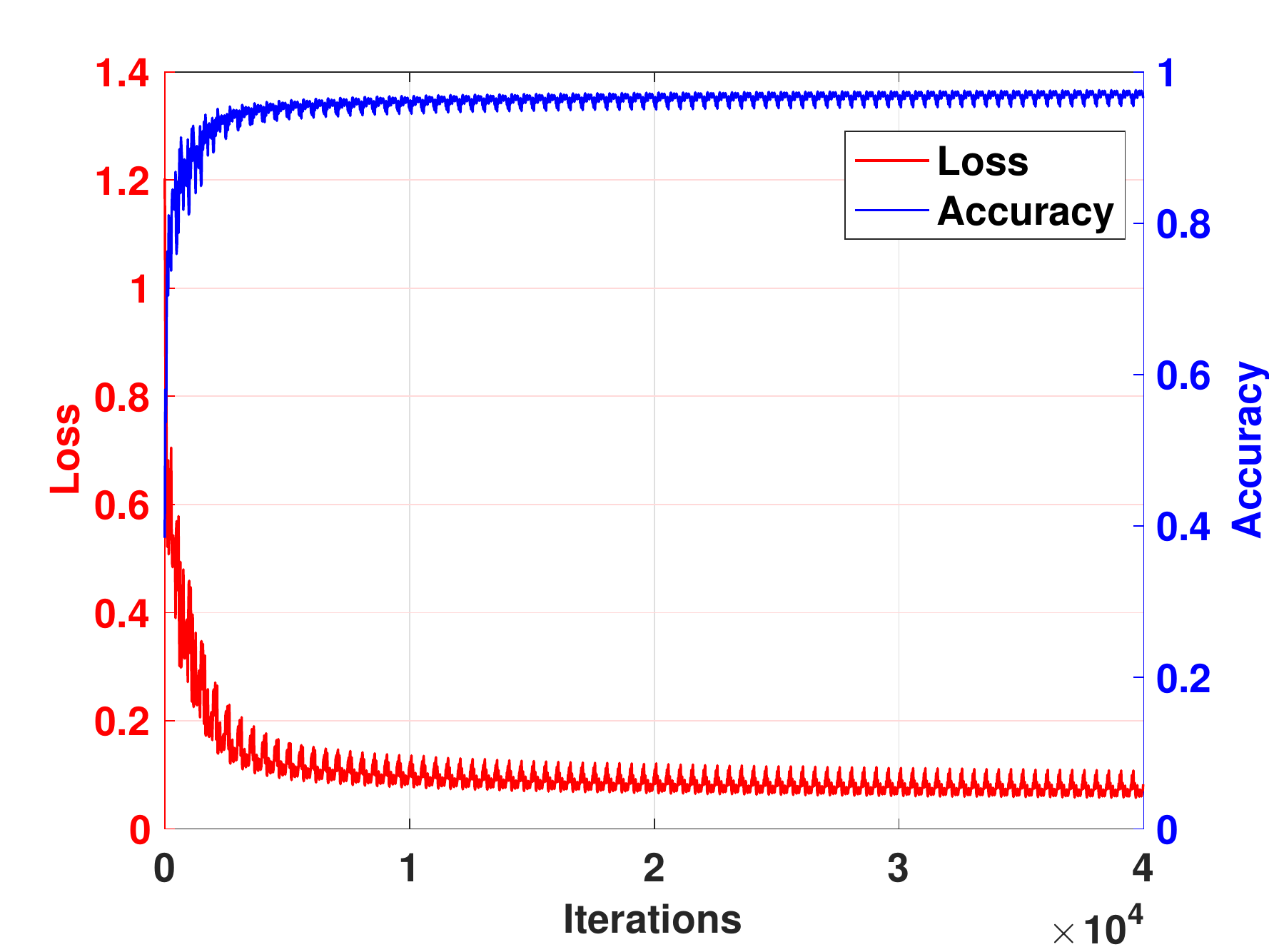}
\end{center}
        \vspace{-4mm}
    \caption{Loss and average class accuracy of three input channels with fine-tuning over various iterations. The maximum number of iterations is set to 40,000, which takes 12\unit{hrs}.}
    \label{fig:loss}
        \vspace{-5mm}
\end{figure}

We use MATLAB to convert the collected datasets to the SegNet data format and annotate the images. A modified version of Caffe~\citep{jia2014caffe} with cuDNN processes input data using CUDA, C++ and Python 2.7. For model training, we set the following parameters: learning rate = 0.001, maximum iterations = 40,000 (640 epochs), batch size = 6, weight delay rate = 0.005 and Stochastic Gradient Descent (SGD) solver~\citep{bottou2010large} is used for the optimization. The average model training time given the maximum number of iterations is 12\unit{hrs}. Fig.~\ref{fig:loss} shows the loss and average class accuracy over 40,000 iterations. This figure suggests that 10,000-20,000 maximum iterations are sufficient since there is a very subtle performance improvement beyond this.

\subsection{Quantitative Results}
For quantitative evaluation, we use a harmonic F1 score measure capturing both precision and recall performance as:
\begin{align}
F1(c)=&2\cdot\frac{\text{precision}_{\mbox{\tiny c}}\cdot \text{recall}_{\mbox{\tiny c}}}{\text{precision}_{\mbox{\tiny c}}+\text{recall}_{\mbox{\tiny c}}}\\ \notag
\text{precision}_{\mbox{\tiny c}}=&\frac{TP_{\mbox{\tiny c}}}{TP_{\mbox{\tiny c}}+FP_{\mbox{\tiny c}}}\\ \notag
\text{recall}_{\mbox{\tiny c}}=&\frac{TP_{\mbox{\tiny c}}}{TP_{\mbox{\tiny c}}+FN_{\mbox{\tiny c}}} \notag
\label{eq:pfh}
\end{align}
where $\text{precision}_{\mbox{\tiny c}}$ and ${TP_{\mbox{\tiny c}}}$ indicate the precision and number of true positives, respectively, for class $c$. The same subscript convention is applied to other notation. Note that the output of SegNet are the probabilities of each pixel belonging to each defined class. In order to compute four fundamental numbers (i.e., TP, TN, FP, and FN), the probability is converted into a binary value. We simply assign the class label of maximum probability and compute precision, recall, accuracy, and F1-score. All models presented in this section are trained and tested with the datasets in Table~\ref{tbl:dataset}.

\begin{figure}
\begin{center}
\includegraphics[width=\columnwidth]{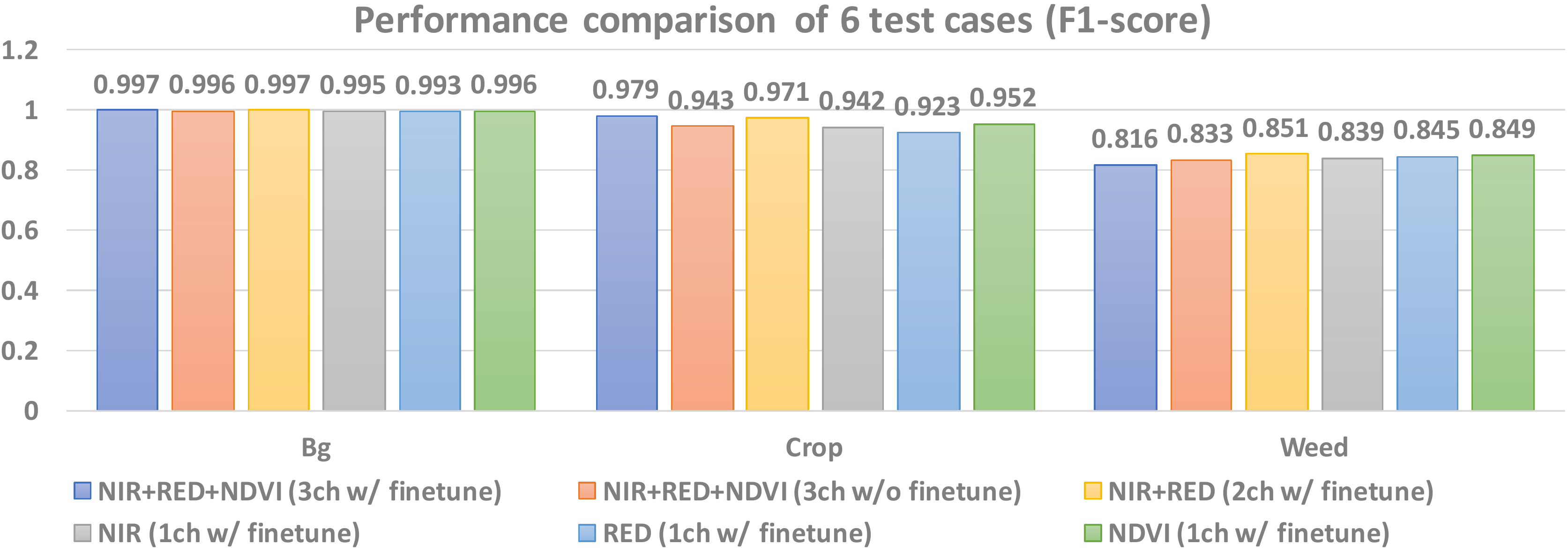}
\end{center}
    \vspace{-2mm}
    \caption{F1-scores of 6 models per classes (horizontal axis). Higher values indicate the better detection performance. Note that the right bars show weed classification F1-score.}
    \label{fig:F1-scores}
        \vspace{-4mm}
\end{figure}

Fig.~\ref{fig:F1-scores} shows the F1-scores of 6 different models trained with varying input data and conditions. There are three classes; \texttt{Bg}, \texttt{Crop}, and \texttt{Weed}.  \texttt{Bg} indicates background (mostly soil but not necessary), and \texttt{Crop} is the sugar beet plant class. All models perform reasonably well (above 80\% for all classes) considering the difficulty of the dataset. In our experiments, we vary two conditions; using a pre-trained model (i.e., VGG16) for network initialization (fine-tuning) and varying the number of input channels

As shown in Fig.~\ref{fig:F1-scores}, the dark blue and orange (with and without fine-tuning given 3 channel input images, respectively) showed that fine-tuning does not impact the output significantly in comparison with more general object detection tasks (e.g., urban scenes or daily life objects). This is mainly because our training datasets (sugar beet and weed images) have different appearance and spectra (660-790\unit{nm}) for the pre-trained model based on RGB ImageNet dataset~\cite{Russakovsky:2015aa}. There may be very few scenes that are similar to our training data. Moreover, the size of dataset we have is relatively small compared with the pre-trained ImageNet (1.2 Million images).

The yellow, gray, light blue, and green bars in Fig.~\ref{fig:F1-scores} present performance measures with varying numbers of input channels. We expected more input data to yield better results since the network captures more useful features that help to distinguish between classes. This can be seen by comparing the results of the 2 channel model (NIR+Red, yellow bar from Fig.~\ref{fig:F1-scores}) and the 1 channel model of NIR and Red (gray and light blue respectively). 2 channel model outperforms for both crop and weed classification. However, interestingly, the number of input data does not always guarantee performance improvement. For instance, the NIR+Red model surpasses the 3 channel model for weed classification performance. We are still investigating this and suspect that it could be due to i) NDVI image is produced based on NIR and Red images meaning that NDVI depends on those two images rather than capturing new information, ii) inaccurate image alignment of NIR and Red image channel on the edge of image where larger distortions exist than in the optical center. Inference performance with varying input data, is discussed in Section~\ref{subsec:tx2}.

\begin{figure}
\begin{center}
\includegraphics[width=0.7\columnwidth]{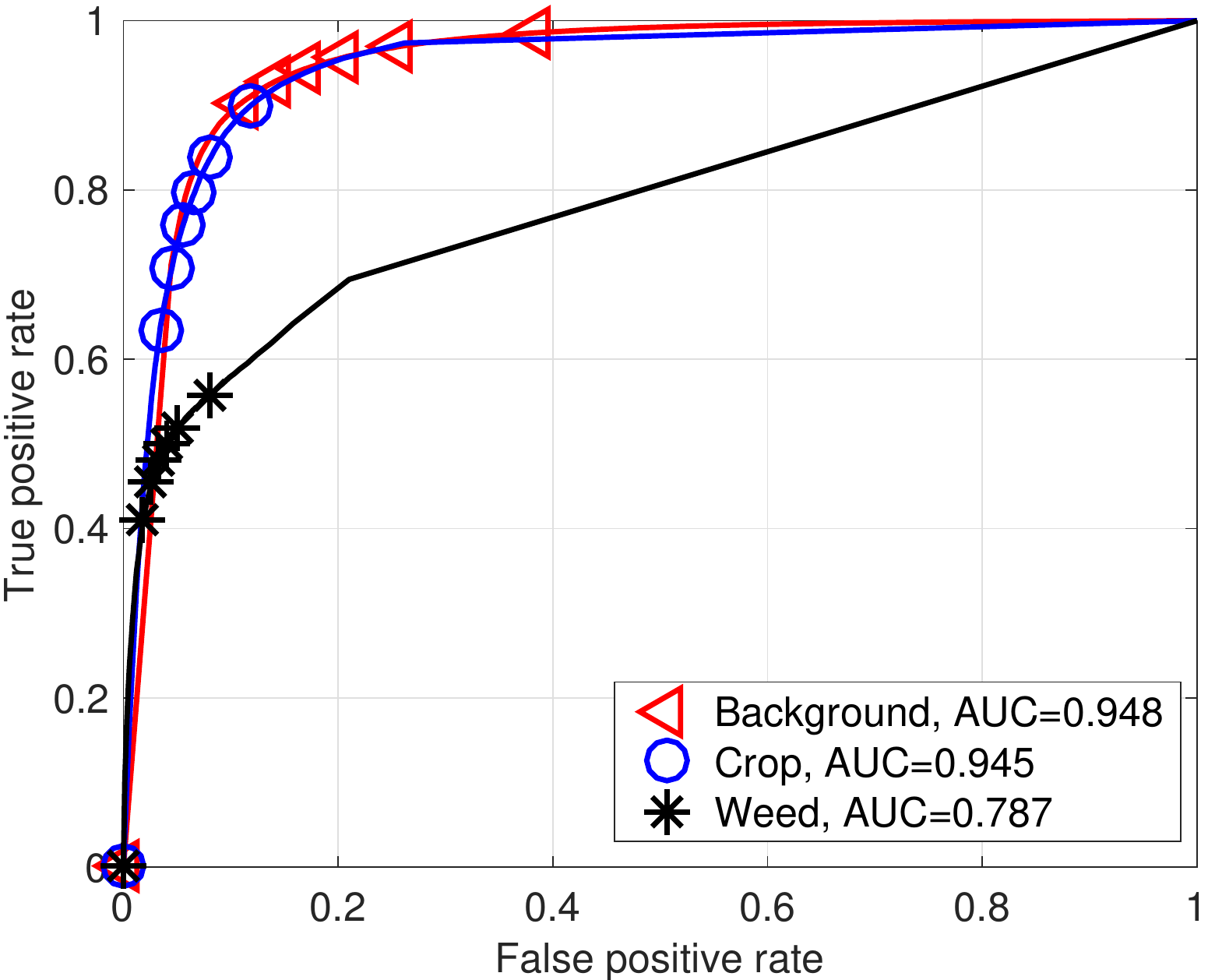}
\end{center}
    \vspace{-3mm}
    \caption{AUC for our 3 channel model corresponding to dark blue bars (left-most) in Fig.~\ref{fig:F1-scores}. Note that the numbers do not match since these measures capture slightly different properties.}
    \label{fig:auc}
        \vspace{-3mm}
\end{figure}

We also use area-under-the-curve (AUC) measures for quantitative evaluation. For example, Fig.~\ref{fig:auc} shows the AUC of 3 channel model. It can be seen that there is small performance variation. As these measures capture different classifier properties, they it cannot be directly compared.

\subsection{Qualitative Results}
\begin{figure*}
\begin{center}
\includegraphics[width=1.6\columnwidth]{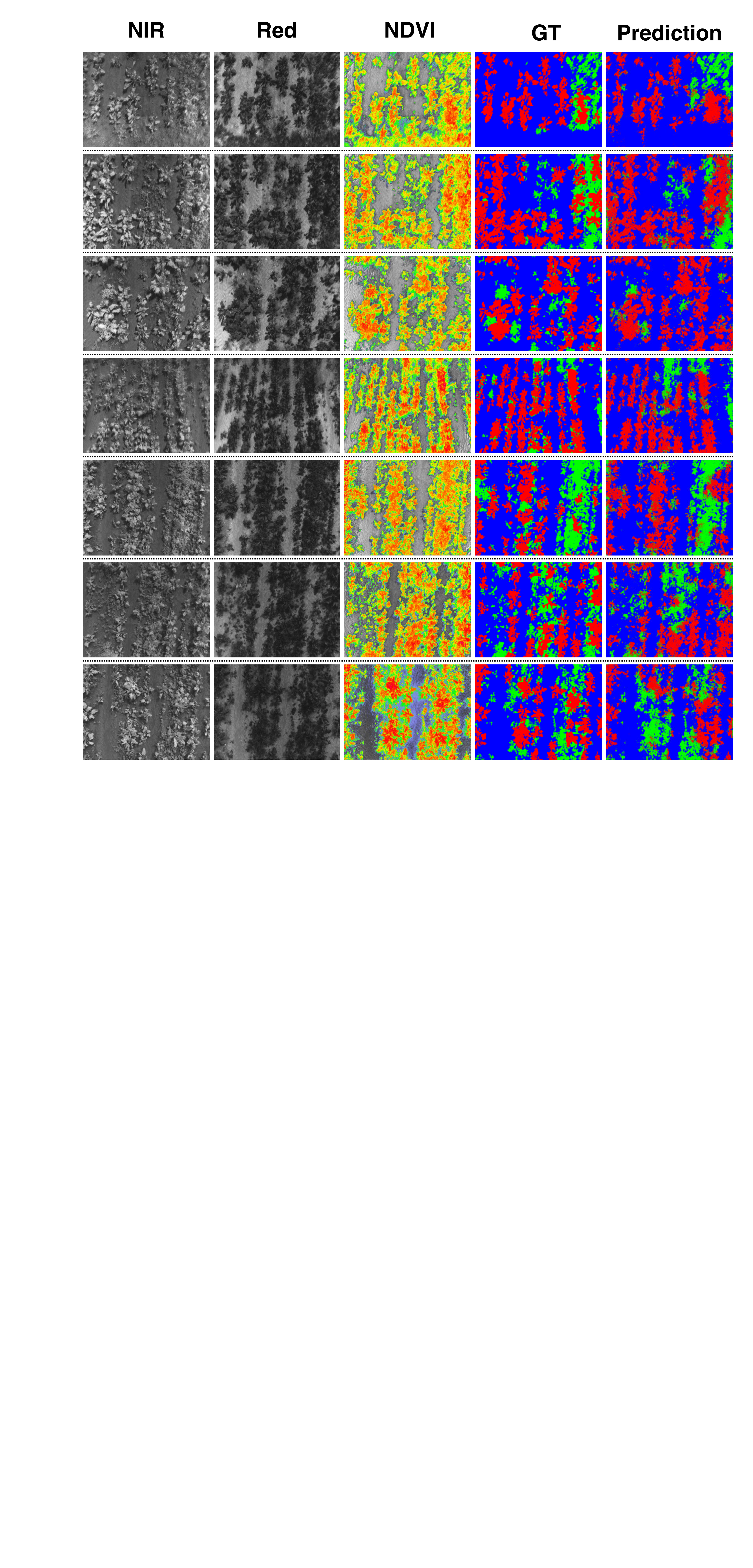}
\end{center}
    \vspace{-4mm}
    \caption{The qualitative results of 7 frames (row-wise). The first three columns are input data to the CNN, with the fourth and fifth showing ground-truths and probability predictions. Because we map the probability of each class to the intensity of R, G, B for visualization, some boundary areas have mixed color.}
    \label{fig:quali}
        \vspace{-4mm}
\end{figure*}

We perform a qualitative evaluation with our best performance model. For visual inspection, we present 7 instances of all input data, ground-truth, and network probability output in Fig.~\ref{fig:quali}. Each row displays an image frame, with the first three columns showing the input of our 3 channel model. NDVI is displayed as heat map scale (colors closer to red depict higher response). The fourth column is annotated ground-truth, and the fifth is our probability output. Note that each class probability is color-coded as intensities of the corresponding color such that background, crop, and weed represent blue, red, and green, respectively. It can be seen that some boundary areas have mixed color due to this. There are noticeable misclassification areas of crop detection in the last row and evident weed misclassification in the second and fifth. This mostly occurs when crops or weeds are surrounded by each other, implying that the network captures not only low-level features, such as edges or intensities, but also object shapes and textures.

\subsection{Discussion, Limitations, and Outlook}
We demonstrated a crop/weed detection pipeline that uses training/testing datasets obtained from consistent environmental conditions. Although it shows applicable qualitative and quantitative results, it is important to validate how well it handles scale variance, and its spatio-temporal consistency. 

\begin{figure}[H]
\begin{center}
\includegraphics[width=\columnwidth]{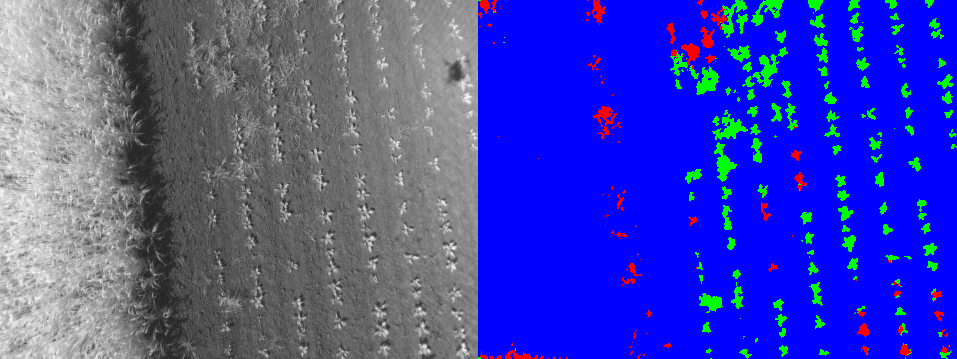}
\end{center}
    \vspace{-3mm}
    \caption{Test image from a different growth stage of sugarbeet plants (left) and its output (right). The same color code is applied for each class as in Fig~\ref{fig:quali}, i.e., blue=\texttt{bg}, red=\texttt{crop}, and green=\texttt{weed}.). This shows that our model reports a high false-positive rate because our training dataset has bigger crops and different type of weeds. Most likely, this is caused by a limited crop/weed temporal training dataset.}
        \vspace{-3mm}
    \label{fig:consisTest}
\end{figure}

We study these aspects using an independent image in Fig.~\ref{fig:consisTest}. The image was taken from a different sugar beet field taken a month prior to the dataset we used (same altitude and sensor). It clearly shows that most of the crops are classified as weed (green) meaning our classifier requires more temporal training dataset. As shown in Fig.~\ref{fig:quali}, we trained a model with larger crops and weeds images than in the new image. Additionally, there may be a different type of weed in the field that the model has not been trained for.

This exemplifies an open issue concerning supervised learning approaches. To address this, we require more training data covering multi-scale, wide-weed varieties over longer time periods to develop a weed detector with spatio-temporal consistency or smarter data augmentation strategies. Even though manually annotating each image would be a labor intensive task, we are planning to incrementally construct a large dataset in the near future. 

\subsection{Inference on an Embedded Platform}\label{subsec:tx2}
Although recent developments in the deep CNN have played a significant role in computer vision and machine learning (especially object detection and semantic feature learning), there is still an issue of running trained models with relatively fast speed (2$\sim$5\unit{Hz}) on a physically constrained device, which can be deployed on a mobile robot. To address this, researchers often utilize a ground station that has a decent GPU computing power with WiFi connection \cite{giusti2016machine}. However, this may cause a large time delay, and it may be difficult to ensure wireless communication between a robot and ground station if coverage is somewhat limited.

\begin{figure}[H]
\begin{center}
\vspace{-4mm}
\subfloat[]{\includegraphics[height=0.34\columnwidth]{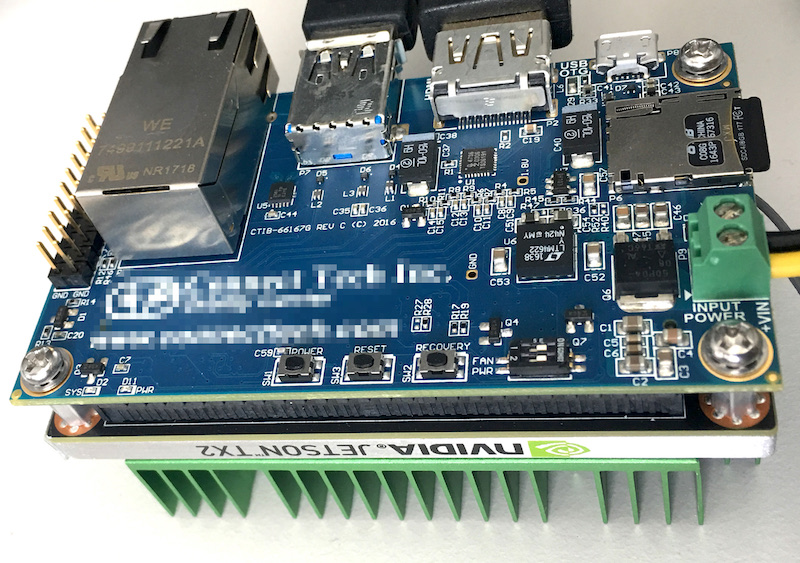}}
\subfloat[]{\includegraphics[height=0.4\columnwidth]{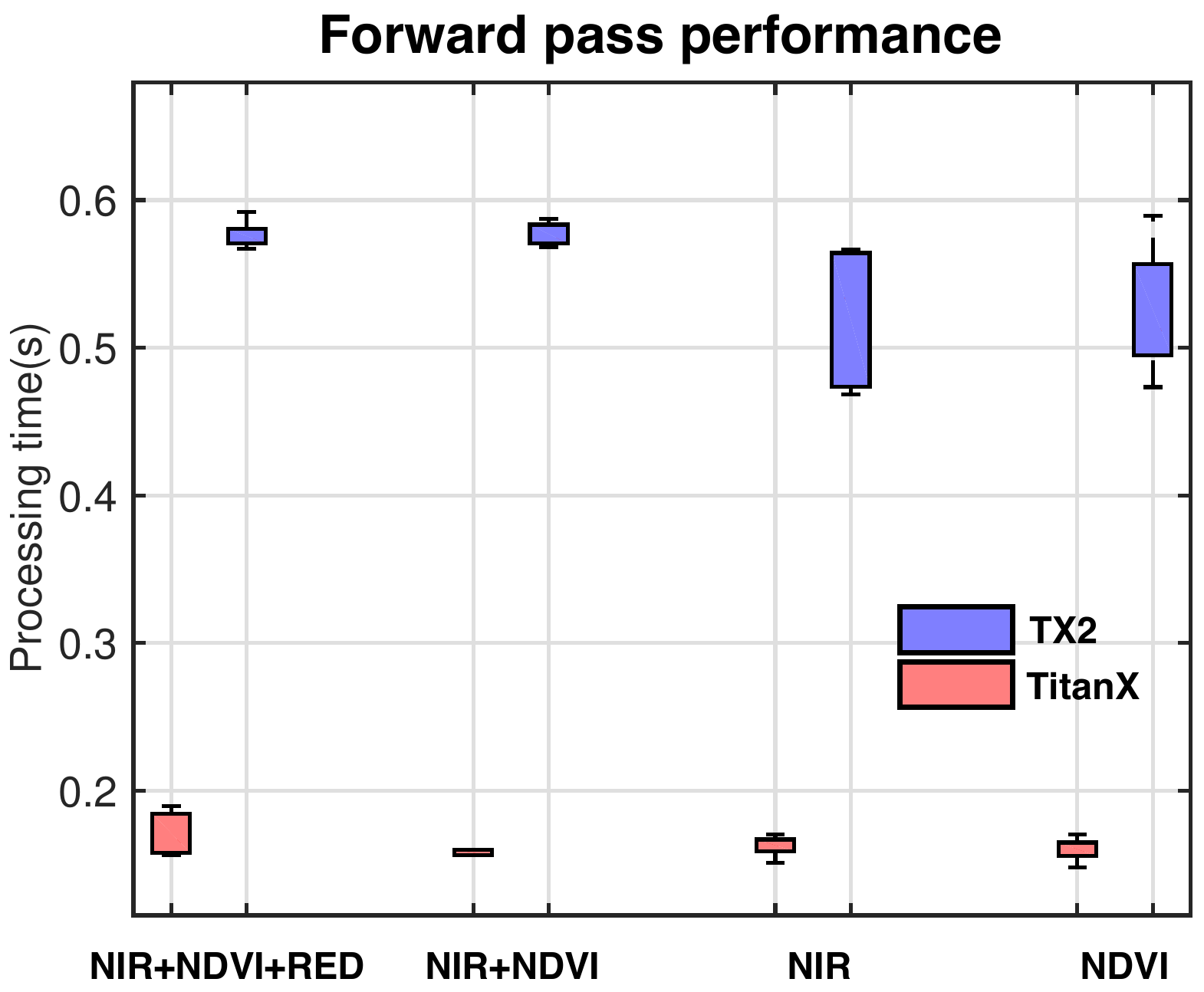}}\newline

\end{center}
\vspace{-4mm}
	\caption{(a) An embedded GPU module (190\unit{g} and 7.5\unit{W}) on which we deploy our trained model. (b) Network forward-pass processing time (y-axis) for different models (x-axis) using Tixan X (red) and Jetson TX2 (blue). (c) Processing time profiling of CPU and GPU of Jetson TX2 while performing the forward-pass.}
	\label{fig:tx2}
	    \vspace{-3mm}
\end{figure}

Using an onboard GPU computer can resolve this issue. Recently an embedded GPU module, Jetson TX2, has been released that performs reasonably well as shown in Fig.~\ref{fig:tx2}(a). It has 2\unit{GHz} hexa-CPU cores, 1.3\unit{GHz} 256 GPU cores, and consumes 7.5\unit{W} while idle and 14\unit{W} for maximum utilization. Fig.~\ref{fig:tx2}(b) shows processing time comparison between Titan X (blue) and TX2 (red). We process 300 images using 4 models denoted in the x-axis. Titan X performs 3.6 times faster than TX2 but considering its power consumption ratio, 17.8 (250\unit{W} for Titan X maximum utilization), TX2 performs significantly well. Another interesting observation in Fig.~\ref{fig:tx2}(b) is that network forward-pass processing time does not affect the number of input channels much. This is because the first multi-convolution layer of all these models has the same filter size of 64. While varying number of inputs affects the contents of these low-level features (e.g., captures different properties of a plant), the size remains identical.


Note that offline inference (i.e., loading images from stored files) is performed on TX2 since the multispectral camera only allows for saving images to a storage device. We are planning to integrate another hyperspectral camera (e.g., Xemia snapshot camera) for online and real-time object detection that can be interfaced with a control~\cite{Kamel:2017aa} or informative path planning~\cite{Popovic2017IROS} modules.
\section{CONCLUSIONS}\label{sec:conclusion}
We demonstrated CNN-based dense semantic classification for weed detection with aerial multispectral images taken from an MAV. The encoder-decoder cascaded deep neural network is trained on a dataset obtained from a herbicide-controlled sugar beet field to address labor intensive labeling tasks. The data obtained from this field is categorized into images containing only crops or weeds, or a crop-weed mixture. For the homogeneous imagery data, vegetation is automatically distinguished by extracting NDVI from multispectral images and applying classic image processing for model training. For the mixed imagery data, we performed manual annotation taking $\sim$ 30 hours.

We trained 6 different models trained on varying numbers of input channels and training conditions and evaluated them quantitatively using F1-scores and AUC as metrics. A qualitative assessment was then performed by a visual comparison of ground-truth with probability prediction outputs. Given the test dataset (mixed), the proposed approach reports an acceptable performance of $\sim$ 0.8 F1-score for weed detection. However, we found spatio-temporal inconsistencies in our model due to limitations in the dataset it was trained on.

We then deploy the model on an embedded system that can be carried by a small MAV, and compare its performance to a high-performance desktop GPU in terms of inference speed and accuracy. Our experimental results estimate that the proposed deep neural network system can run the high-level perception task at 1.8\unit{Hz} on the embedded platform which can be deployed on a MAV.

Finally, multispectral weed and crop images with the corresponding ground truth used in this paper are released for the robotics community and to support future work in agricultural robotics domain.
\section{Acknowledgement}\label{sec:ack}
\vspace{-2mm}
This project has received funding from the European Union's Horizon 2020 research and innovation programme
under grant agreement No 644227, No 644128 and from the Swiss State Secretariat for Education, Research
and Innovation (SERI) under contract number 15.0029 and 15.0044.

We also gratefully acknowledge the support of NVIDIA Corporation with the donation of the Titan X Pascal GPU used for this research and Hansueli Zellweger for the management of the experimental field.


\bibliographystyle{IEEEtranN}
\footnotesize
\bibliography{bibs/RAL2018}

\end{document}